\pdfoutput=1

\documentclass[11pt]{article}

\usepackage[preprint]{acl}

\usepackage{times}
\usepackage{latexsym}

\usepackage[T1]{fontenc}

\usepackage[utf8]{inputenc}

\usepackage{microtype}

\usepackage{inconsolata}

\usepackage{tabularx} 
\usepackage{arydshln}
\usepackage{amsmath}
\usepackage{diagbox}
\usepackage{caption}
\usepackage{amsfonts,amssymb}
\usepackage{subfigure}
\usepackage{graphicx}
\usepackage{bm}
\usepackage{multirow}
\usepackage[ruled,lined,commentsnumbered]{algorithm2e}
\usepackage{hyperref}
\usepackage{booktabs}
\usepackage{bbm}
\usepackage[inkscapelatex=false]{svg}
\usepackage{pifont}
\usepackage{stfloats}
\usepackage{multicol}
\usepackage{float}
\usepackage{array}
\usepackage{amssymb}
\newcommand{\PreserveBackslash}[1]{\let\temp=\\#1\let\\=\temp}
\newcolumntype{C}[1]{>{\PreserveBackslash\centering}p{#1}}
\newcolumntype{R}[1]{>{\PreserveBackslash\raggedleft}p{#1}}
\newcolumntype{L}[1]{>{\PreserveBackslash\raggedright}p{#1}}

\usepackage{mathrsfs}

\title{Improving Multilingual Neural Machine Translation by Utilizing Semantic and Linguistic Features}

\author{
    Mengyu Bu\textsuperscript{\rm 1,3},
    Shuhao Gu\textsuperscript{\rm 1,3}\footnotemark[2],
    Yang Feng\textsuperscript{\rm 1,2,3}\footnotemark[1] \\
    \textsuperscript{\rm 1}{Key Laboratory of Intelligent Information Processing,} \\ Institute of Computing Technology, Chinese Academy of Sciences (ICT/CAS) \\
    { \textsuperscript{\rm 2} {Key Laboratory of AI Safety, Chinese Academy of Sciences}} \\
    { \textsuperscript{\rm 3} {University of Chinese Academy of Sciences, Beijing, China}} \\
     \texttt{\{\href{mailto:bumengyu23z@ict.ac.cn}{bumengyu23z}, \href{mailto:fengyang@ict.ac.cn}{fengyang}\}@ict.ac.cn, \href{mailto:shuhaog515@gmail.com}{shuhaog515@gmail.com}}}

\begin{document}
\maketitle

\renewcommand{\thefootnote}{\fnsymbol{footnote}} 
\footnotetext[1]{Corresponding author: Yang Feng.} 
\footnotetext[2]{This paper was done when Shuhao Gu studied at ICT/CAS.} 
\renewcommand{\thefootnote}{\arabic{footnote}}

\begin{abstract}

The many-to-many multilingual neural machine translation can be regarded as the process of integrating semantic features from the source sentences and linguistic features from the target sentences.
To enhance zero-shot translation, models need to share knowledge across languages, which can be achieved through auxiliary tasks for learning a universal representation or cross-lingual mapping.
To this end, we propose to exploit both semantic and linguistic features between multiple languages to enhance multilingual translation.
On the encoder side, we introduce a disentangling learning task that aligns encoder representations by disentangling semantic and linguistic features, thus facilitating knowledge transfer while preserving complete information.
On the decoder side, we leverage a linguistic encoder to integrate low-level linguistic features to assist in the target language generation.
Experimental results on multilingual datasets demonstrate significant improvement in zero-shot translation compared to the baseline system, while maintaining performance in supervised translation.
Further analysis validates the effectiveness of our method in leveraging both semantic and linguistic features.\footnote{The code is available at \url{https://github.com/ictnlp/SemLing-MNMT}.}

\end{abstract}
\section{Introduction}

The many-to-many multilingual neural machine translation (NMT) enables translation across multiple languages in a single model \citep{firat-etal-2016-multi, fan2021beyond, siddhant-etal-2020-leveraging}. 
This process involves integrating the semantic information from the source sentence and the linguistic features from the target sentence.
Specifically, the encoder captures the semantic information of the source sentence and maps it to the representation space, when the decoder integrates this representation with the inherent features of the target language to generate the target language sentences. 
By sharing parameters, multilingual NMT models enable knowledge transfer across languages, which significantly benefits low-resource machine translation~\cite{aharoni-etal-2019-massively}, especially zero-shot translation. 
However, previous multilingual NMT models do not explicitly differentiate between semantic and linguistic features, resulting in the entanglement of knowledge and linguistics within the model. 
This entanglement negatively affects the performance of zero-shot translation in two main aspects. 
Firstly, the linguistic features of the source language interfere with the encoder's ability to learn shared semantic information, causing the encoder to encode different languages into different representation subspaces \cite{kudugunta-etal-2019-investigating}, which hinders knowledge transfer across multiple languages. 
Secondly, during the generation of the target language, the decoder is influenced by spurious correlations due to insufficient target language features, leading to off-target issues \citep{zhang-etal-2020-improving}.

Numerous methods have been proposed to address these challenges. 
Some methods focus on learning a universal cross-lingual representation space at the encoder \citep{arivazhagan2019missing, pan-etal-2021-contrastive, pham-etal-2019-improving, gu-feng-2022-improving}, while others aim to facilitate target language generation at the decoder \citep{yang-etal-2021-improving-multilingual, gao-etal-2023-improving}.
For example, \citet{pan-etal-2021-contrastive} employ contrastive learning to minimize the distance between parallel sentence pairs and maximize the distance between irrelevant sentence pairs. \citet{gu-feng-2022-improving} leverage optimal transport theory to learn a universal representation and introduce an agreement-based training approach to make consistent predictions. \citet{gao-etal-2023-improving} promote consistent cross-lingual representation by cross-lingual consistency regularization.
However, these methods primarily concentrate on learning language-agnostic semantic features while neglecting the language-specific linguistic features. Such incompleteness of information can lead to less accurate semantic mapping and alignment degradation, which can harm the supervised and zero-shot translation performance.

\begin{figure*}[htbp]
  \centering
  \includegraphics[width=0.85\linewidth]{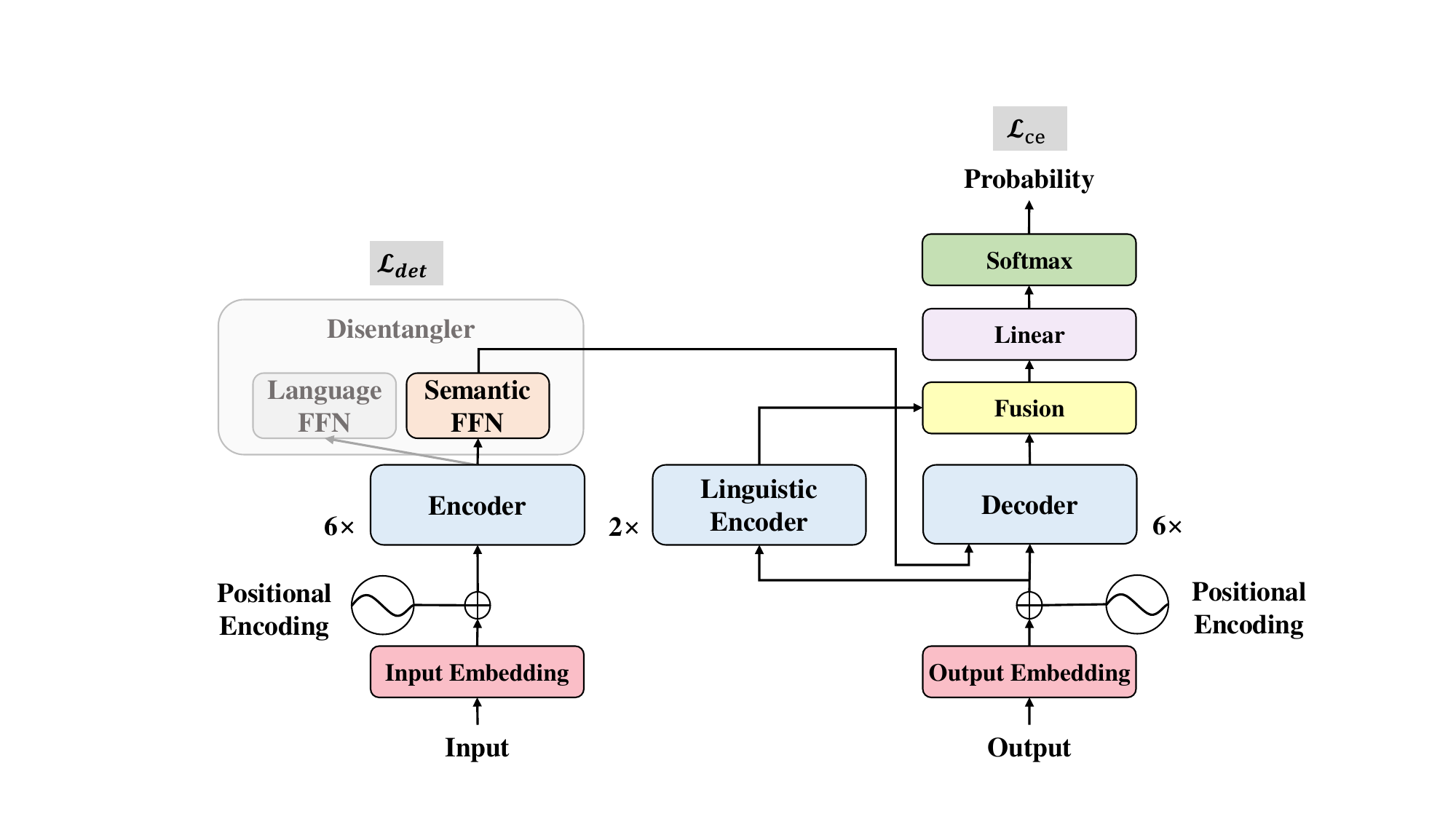}
  \caption{The framework of our method. We propose a disentangler to learn a universal semantic representation via disentangling and utilize a linguistic encoder to fuse low-level linguistic features.}
  \label{model_arch}
\end{figure*}

In this paper, we propose to utilize both semantic and linguistic features for multilingual NMT.
Our insight is to learn a semantic representation space non-destructively and utilize linguistic features to guide target language generation, thereby improving the performance of zero-shot translation and maintaining supervised translation performance lossless.
Specifically, at the encoder, we introduce a disentangler and disentangling learning task to facilitate harmless cross-lingual semantic alignment.
At the decoder, we utilize a linguistic encoder to integrate low-level linguistic features from lower layers and high-level semantic features from higher layers. This fusion provides implicit guidance for the target language generation.

Experiments on multilingual datasets show that our method brings an average of 0.18+ BLEU in the supervised translation direction and an average of 3.74+ BLEU in the zero-shot translation direction compared with the baseline system. Furthermore, the analysis demonstrates that our method can alleviate the problems caused by the entanglement of semantic and linguistic features, obtain a better semantic representation, and reduce the off-target rate in zero-shot translation.

\section{Background}

In this section, we will give a brief introduction to the Transformer \citep{vaswani2017attention} model for NMT and many-to-many multilingual NMT.

\subsection{Transformer for NMT}

Define the set $\mathcal{D}$ of parallel corpus with input sentence $\mathbf{x}_i^u$ and reference $\mathbf{y}_i^v$, where $u$ and $v$ denote language type and $i$ denotes the sentence id, which is the same for a sentence pair. The model first maps the source sentence $\mathbf{x}_i^u$ from the token representation to the token vector representation $embed(\mathbf{x}_i^u)$. After that, the encoder encodes the token vector sequence into the hidden state representation $\mathbf{h}(\mathbf{x}_i^u)$. Similarly, the model maps the predicted $k-1$ target words into a sequence of token vectors, and the decoder decodes the $k$th target word based on the sequence of hidden states $\mathbf{h}(\mathbf{x}_i^u)$ from the encoder and the predicted sequence of $k-1$ target words. The model repeats this process until the $\langle eos \rangle$ token is predicted. The model is optimized by minimizing the cross-entropy (CE) loss between the predicted sentence and the reference as follows:
\begin{equation}
    \mathcal{L}_{ce} = -\sum_{\mathbf{x}_i^u, \mathbf{y}_i^v \in \mathcal{D}} \log (P_\theta (\mathbf{y}_i^v | \mathbf{x}_i^u))
\end{equation}
where $\theta$ denotes the model parameters.

\subsection{Many-to-Many Multilingual NMT}

The many-to-many multilingual NMT model supports translation between multiple languages. Formally, we define the set of languages $L=\{L_1,...,L_M\}$, where $M$ is the number of languages. We prepend a language token at the beginning of the source sentence and the target sentence to indicate the language type respectively. For example, the following sentence pair translated from English to French: "Hello world! $\to$ Bonjour le monde!" is transformed into "$\langle en \rangle$ Hello world! $\to$ $\langle fr \rangle$ Bonjour le monde!".

\section{Method}

The core idea of our approach is to improve multilingual NMT by exploiting semantic and linguistic features at the same time. To achieve this, we disentangle semantic and linguistic features for the encoder and utilize low-level linguistic features for the decoder. Specifically, we propose a disentangler at the encoder to facilitate lossless semantic alignment. Meanwhile, we introduce a linguistic encoder at the decoder to fuse low-level features to guide target language generation. Figure \ref{model_arch} illustrates the framework of our approach.

\begin{figure}[t!]
  \centering
  \includegraphics[width=1.0\linewidth]{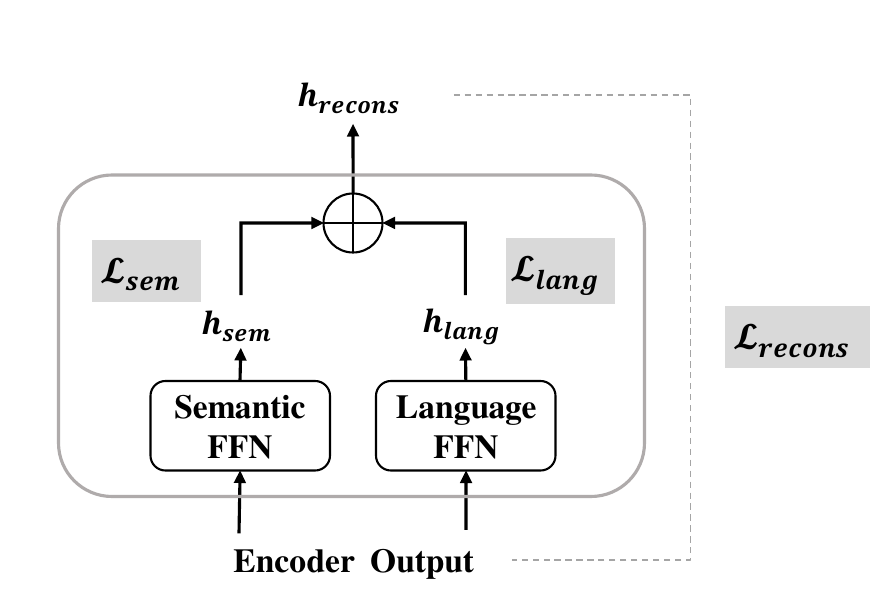}
  \caption{The architecture of disentangler. $\bigoplus$ denotes the summation of the feature dimensions.}
  \label{disentangler_arch}
\end{figure}

\subsection{Disentangling Semantic and Linguistic Features}

Empirically, sentences possess both semantic and linguistic properties, which can be expressed as "sentence = semantics + linguistics". Semantic features encompass the meanings of language units that are shared across languages, while linguistic features comprise the rules for constructing sentences, such as lexical labels and morphological information. By disentangling linguistic features from sentences, we can obtain a universal semantic representation shared by different languages.
Referring to the design of \citet{tiyajamorn-etal-2021-language} on multilingual language understanding tasks, we extract semantic and linguistic features separately and implement interactive decomposition using reconstruction constraints.

The structure of our disentangler is shown in Figure \ref{disentangler_arch}. The disentangler consists of two feed-forward networks (FFN), a semantic FFN and a language FFN, respectively. For the encoder representations, the former extracts language-agnostic semantic information, and the latter extracts language-specific linguistic features. Then, the two outputs are summed to reconstruct the original encoder representations. The extracted semantic features are fed into the decoder.

We define disentangling loss $\mathcal{L}_{det}$ to optimize disentangler by multi-task learning:
\begin{equation}
    \mathcal{L}_{det} = \mathcal{L}_{sem} + \mathcal{L}_{lang} + \lambda_1 \mathcal{L}_{recons}
\end{equation}
where $\mathcal{L}_{sem}$ assists the semantic FFN to extract semantic information, $\mathcal{L}_{lang}$ assists the language FFN to extract linguistic features, $\mathcal{L}_{recons}$ is the reconstruction constraint, and $\lambda_1$ is the hyperparameter to balance the losses.

\paragraph{Semantic Learning} To learn a universal representation space across multiple languages, we explicitly minimize the distance of parallel sentence pairs and maximize the distance of irrelevant sentence pairs. For the multilingual parallel dataset $\mathcal{D}$, we make the parallel sentence pair $(\mathbf{x}_i^u,\mathbf{y}_i^v ) \in \mathcal{D}$ a positive example and randomly choose a sentence $\mathbf{y}_j^w$ from language $L_w$ as a negative example $(\mathbf{x}_i^u,\mathbf{y}_j^w )$, where $L_u$ can be the same as $L_w$. We optimize by minimizing the following loss function:
\begin{equation}
    \begin{aligned}
        \mathcal{L}_{sem} = &\mathbb{E}_{\mathbf{x}_i^u,\mathbf{y}_i^v \in \mathcal{D}} (1 - sim^+(\mathcal{R}(\mathbf{x}_i^u), \mathcal{R}(\mathbf{y}_i^v)) + \\
        &\lambda_2 \mathbb{E}_{\mathbf{x}_i^u,\mathbf{y}_j^w \in \mathcal{D}} (1 + sim^-(\mathcal{R}(\mathbf{x}_i^u), \mathcal{R}(\mathbf{y}_j^w)))
    \end{aligned}
\end{equation}
where $sim(\cdot)$ calculates the similarity of different sentences, and we use cosine similarity to evaluate the similarity. $+$ and $-$ denote positive and negative samples, respectively. $\mathcal{R}(\cdot)$ calculates the average-pooled hidden state representation. $\lambda_2$ is the hyperparameter that balances positive and negative samples. To simplify the implementation in the training, we sample the positive and negative samples within every training batch.

\paragraph{Language Learning} Empirically, different languages have different linguistic features, such as lexical and syntactic patterns. Compared with sentences in different languages, the linguistic features are more similar between sentences in the same language. To extract linguistic features of multiple languages, we design the training objective similar to semantic learning. We minimize the distance of sentences in the same language and maximize the distance of sentences in different languages. We make the same language sentence pair $(\mathbf{x}_i^u,\mathbf{y}_j^u) \in \mathcal{D}$  a positive example and choose a random sentence $\mathbf{y}_k^v$ from any other language $L_k$ as a negative example $(\mathbf{x}_i^u, \mathbf{y}_k^v)$. We optimize by minimizing the following loss function:
\begin{equation}
    \begin{aligned}
        \mathcal{L}_{lang} = &\mathbb{E}_{\mathbf{x}_i^u,\mathbf{y}_j^u \in \mathcal{D}} (1 - sim^+(\mathcal{R}(\mathbf{x}_i^u), \mathcal{R}(\mathbf{y}_j^u)) + \\
        &\lambda_2 \mathbb{E}_{\mathbf{x}_i^u,\mathbf{y}_k^v \in \mathcal{D}} (1 + sim^-(\mathcal{R}(\mathbf{x}_i^u), \mathcal{R}(\mathbf{y}_k^v)))
    \end{aligned}
\end{equation}
Where $sim(\cdot)$, $+$, $-$ and $\mathcal{R}(\cdot)$ are defined the same as semantic learning. $\lambda_2$ is a shared hyperparameter. We sample the positive and negative samples within every training batch.

\paragraph{Reconstruction Constraint} 
To make the semantic and linguistic features fully interact and learn a more reasonable semantic representation, we design the reconstruction constraint to utilize the linguistic features. We input the encoded hidden state $\mathbf{h}(x_i^u)$ from the encoder into the disentangler, extract the semantic information $\mathbf{h}_{sem} (x_i^u)$ and the language features $\mathbf{h}_{lang} (x_i^u)$ respectively, and reconstruct the original encoder representation by summation. We perform optimization by minimizing the following loss function:
\begin{equation}
    \begin{aligned}
        \mathcal{L}_{recons} = &\mathbb{E}_{\mathbf{x}_i^u \in \mathcal{D}} (\frac{1}{d}||\mathbf{h}(\mathbf{x}_i^u)) - \\
        & \mathbf{h}_{sem}(\mathbf{x}_i^u) - \mathbf{h}_{lang}(\mathbf{x}_i^u)||_2)
    \end{aligned}
\end{equation}
Where $d$ denotes the average sequence length.

\subsection{Integrating Linguistic Features}

In multilingual NMT models, decoders need to map the semantic representation to the target language representation. For zero-shot translation, since the model has not seen the translation direction during training, it is difficult to model the mapping relationships well, leading to the off-target phenomenon \citep{gu-etal-2019-improved}.
Previous works show that the off-target rate is strongly correlated with the zero-shot translation performance \citep{zhang-etal-2020-improving, wu-etal-2021-language, wang2022understanding}. Therefore, we hope to reduce the off-target rate by fusing additional information to guide the decoder to generate the target language.

\citet{zhang2021share} show that the lower encoder layers will focus more on linguistic information, which is crucial to distinguishing different languages.
Therefore, we introduce a linguistic encoder with two encoder layers, which has the same input as the decoder. We utilize a fusion layer to integrate the low-level representations containing more linguistic features and the high-level representations containing more semantic features. We denote the linguistic encoder output as $\mathbf{h}_{lingEnc}$ and the decoder output as $\mathbf{h}_{dec}$. We concatenate these features in the feature dimension and utilize a two-layer FFN for fusion:
\begin{equation}
    \mathbf{h}=\mbox{W}_1(\mbox{ReLU}(\mbox{W}_2([\mathbf{h}_{dec};\mathbf{h}_{lingEnc}])))
\end{equation}
Where $\mbox{W}_1 \in \mathbb{R}^{d \times d_h}$ and $\mbox{W}_2 \in \mathbb{R}^{2d_h \times d}$ are two feed-forward layers.

\subsection{Joint Training Strategy}

We optimize the final loss function using a joint training strategy, including cross-entropy loss and disentangling loss:
\begin{equation}
    \mathcal{L} = \mathcal{L}_{ce} + \lambda d \mathcal{L}_{det}
\end{equation}
where $\lambda$ is the hyperparameter that controls the contribution of $\mathcal{L}_{det}$. Since $\mathcal{L}_{ce}$ is calculated at the token level and $\mathcal{L}_{det}$ is calculated at the sentence level, we multiply $\mathcal{L}_{det}$ by the average sequence length $d$.

\section{Experiment}

We evaluate the performance of our model on the multilingual benchmarks, including supervised and zero-shot translation performance.

\subsection{Dataset Description}

We conduct experiments on the IWSLT2017, OPUS-7, and PC-6 datasets. The brief information about these datasets is shown in Appendix \ref{sec:dataset_description}.

\paragraph{IWSLT2017} The IWSLT2017 benchmark \citep{cettolo2017overview} contains five languages: German (De), English (En), Dutch (Nl), Romanian (Ro), and Italian (It). We experiment on the English-centric dataset consisting of eight supervised translation directions and twelve zero-shot translation directions. The training set contains 0.22M$\sim$0.26M parallel sentence pairs per language direction. We use the official validation set and test set.

\paragraph{OPUS-7} The OPUS-7 dataset is a subset of the OPUS-100 dataset \citep{pan-etal-2021-contrastive}. The OPUS-7 dataset is English-centric and contains seven languages: Arabic (Ar), German (De), English (En), French (Fr), Dutch (Nl), Russian (Ru), and Chinese (Zh). This dataset contains twelve supervised translation directions and thirty zero-shot translation directions. The training set contains 1M parallel sentence pairs per language direction. We use the official validation set and test set.

\paragraph{PC-6} We construct the PC-6 dataset following \cite{gu-feng-2022-improving}. PC-6 is a dataset of six languages extracted from PC32 \cite{pan-etal-2021-contrastive}, including  Czech (Cs), Kazakh (Kk), Romanian (Ro), Russian (Ru), and Turkish (Tr). The PC-6 dataset contains ten supervised translation directions and twenty zero-shot translation directions. The amount of data in each language direction varies in size from 0.12M to 1.84M sentence pairs. For the supervised translation, we use the official WMT2016$\sim$2019 validation set and test set. For the zero-shot translation, we extract from WikiMatrix \cite{schwenk-etal-2021-wikimatrix} to obtain the test set.

We directly apply the Unigram Model algorithm to preprocess the original multilingual corpus using \textit{Sentencepiece} toolkit~\cite{KudoR18}. We build a shared dictionary across multiple languages, which contains 32K tokens.

\subsection{Model Configuration}

We use the base Transformer configuration~\cite{vaswani2017attention} for both the encoder and decoder. For the disentangler and the fusion layer, we set $ffn\_dim=2048$. For the linguistic encoder, we apply two encoder layers of the base Transformer configuration. We use the Adam optimizer and set $\beta=(0.9, 0.98)$. We set $warmup=4000$ and use the inverse square root learning scheduler with the learning rate set to $7e^{-4}$. We apply label-smoothed cross-entropy loss with a smoothing rate of $0.1$. We set $dropout=0.3$ for the IWSLT2017 dataset and $dropout=0.1$ for the OPUS-7 and PC-6. We set $\lambda = 0.05$, $\lambda_1 = 0.2$ and $\lambda_2 = 0.2$ for our system. We train the models on 4 RTX3090 GPUs. 

In the inference stage, we use beam search with beam size 5 and set the length penalty to 1. We report the BLEU~\cite{PapineniRWZ02}, ChrF~\cite{popovic-2015-chrf} and COMET~\cite{rei-etal-2022-comet}\footnote{\url{https://huggingface.co/Unbabel/wmt22-comet-da}} scores for comprehensive evaluation. The BLEU and ChrF scores are computed using the \textit{SacreBLEU} toolkit~\cite{post-2018-call}. We select the best-performing three checkpoints on the validation sets from the last six checkpoints and report the average scores on the test sets.

\begin{table*}[t]
    \centering
    
    \resizebox{\linewidth}{!}{\begin{tabular}{c|ccccccc|c} \hline
    \multirow{2}*{IWSLT2017} & \multirow{2}*{De $\leftrightarrow$ It} & \multirow{2}*{De $\leftrightarrow$ Nl} & \multirow{2}*{De $\leftrightarrow$ Ro} & \multirow{2}*{It $\leftrightarrow$ Nl} & \multirow{2}*{It $\leftrightarrow$ Ro} &\multirow{2}*{Nl $\leftrightarrow$ Ro} & Zero-shot & Supervised \\
    ~ & ~ & ~ & ~ & ~ & ~ & ~ & Average & Average \\ \hline
    m-Transformer & 16.94 & 20.89 & 17.02 & 17.58 & 17.32 & 17.83 & 17.93 & \textbf{34.32} \\
    mRASP2 w/o AA & 21.39 & 24.67 & 21.40 & 22.43 & \textbf{22.70} & 21.87 & 22.41 & 34.06 \\
    TLP & 19.52 & 23.41 & 19.08 & 20.75 & 20.34 & 20.29 & 20.56 & 34.29 \\
    DT & 18.76 & 22.05 & 18.65 & 20.04 & 20.31 & 19.83 & 19.94 & 34.17 \\
    CrossConST & 20.93 & 24.10 & 20.65 & 21.98 & 21.68 & 21.40 & 21.73 & 34.10 \\ \hdashline
    Pivot & 21.80 & 24.14 & 21.23 & 22.63 & 22.19 & 22.27 & 22.37 & -  \\ \hdashline
    Alpaca & 14.50 & 17.94 & 12.88 & 14.46 & 13.44 & 12.18 & 14.23 & 27.37 \\
    ChatGPT & 23.85 & 27.01 & 23.21 & 25.17 & 24.72 & 24.83 & 24.79 & 35.26 \\ \hdashline
    Ours & \textbf{21.55} & \textbf{25.18} & \textbf{21.50} & \textbf{22.84} & 22.66 & \textbf{22.46} & \textbf{22.70} & 34.25 \\ \hline
    \end{tabular}}
    
    \vspace{0.5em}
    
    \resizebox{\linewidth}{!}{\begin{tabular}{c|ccccccc|c} \hline
    \multirow{2}*{OPUS-7} & \multirow{2}*{x $\rightarrow$ Ar} & \multirow{2}*{x $\rightarrow$ De} & \multirow{2}*{x $\rightarrow$ Fr} & \multirow{2}*{x $\rightarrow$ Nl} & \multirow{2}*{x $\rightarrow$ Ru} &\multirow{2}*{x $\rightarrow$ Zh} & Zero-shot & Supervised \\
    ~ & ~ & ~ & ~ & ~ & ~ & ~ & Average & Average \\ \hline
    m-Transformer & 12.73 & 14.58 & 23.11 & 15.82 & 17.24 & 16.44 & 16.65 & 31.91 \\
    mRASP2 w/o AA & 14.20 & 16.54 & 25.67 & \textbf{17.46} & 19.29 & 17.72 & 18.48 & 31.48 \\
    TLP & 13.10 & 14.32 & 22.71 & 15.82 & 17.29 & 16.29 & 16.59 & 31.90 \\
    DT & 13.84 & 15.06 & 23.95 & 16.24 & 18.29 & \textbf{17.85} & 17.54 & 31.96 \\
    CrossConST & 13.68 & 15.04 & 23.67 & 16.50 & 17.67 & 16.03 & 17.10 & 32.12 \\ \hdashline
    Pivot & 15.79 & 17.48 & 27.81 & 18.50 & 20.98 & 15.71 & 19.38 & - \\ \hdashline
    Ours & \textbf{14.79} & \textbf{16.64} & \textbf{26.20} & 17.42 & \textbf{20.11} & 17.40 & \textbf{18.76} & \textbf{32.13} \\ \hline
    \end{tabular}}
    
    \vspace{0.5em}
    
    \resizebox{\linewidth}{!}{\begin{tabular}{c|cccccc|c} \hline
    \multirow{2}*{PC-6} & \multirow{2}*{x $\rightarrow$ Cs} & \multirow{2}*{x $\rightarrow$ Kk} & \multirow{2}*{x $\rightarrow$ Ro} & \multirow{2}*{x $\rightarrow$ Ru} & \multirow{2}*{x $\rightarrow$ Tr} & Zero-shot & Supervised \\
    ~ & ~ & ~ & ~ & ~ & ~ & Average & Average \\ \hline
    m-Transformer & 8.05 & 4.27 & 11.05 & 10.91 & 5.45 & 7.94 & 20.38 \\
    mRASP2 w/o AA & \textbf{14.27} & 2.85 & 15.67 & 16.12 & 8.22 & 11.42 & 19.93 \\
    TLP & 9.67 & 4.27 & 11.35 & 10.79 & 6.17 & 8.45 & 20.68 \\
    DT & 12.57 & \textbf{4.85} & 14.32 & 15.63 & 7.78 & 11.03 & 20.22 \\
    CrossConST & 13.03 & 2.29 & 14.56 & 13.41 & 7.41 & 10.14 & 20.50 \\ \hdashline
    Pivot & 13.73 & 2.13 & 15.91 & 15.20 & 8.30 & 11.05 & - \\ \hdashline
    Ours & 14.13 & 4.20 & \textbf{16.92} & \textbf{16.72} & \textbf{9.44} & \textbf{12.28} & \textbf{20.77} \\ \hline
    \end{tabular}}
    \caption{Overall performance on the multilingual test sets. We report the BLEU scores. "Zero-shot Average" and "Supervised Average" denote the average BLEU scores on the zero-shot and supervised directions. The "x" in the last two tables denotes all languages except for the target language. We bold the highest BLEU scores except for Alpaca, ChatGPT and Pivot.}
    \label{main_result}
\end{table*}

\subsection{Contrast Systems}

We compare our model with the following multilingual NMT systems. We prepend a language token at the beginning of the sentences for NMT systems.

\paragraph{m-Transformer}~\cite{JohnsonSLKWCTVW17} This method utilizes cross-entropy loss to train a Transformer model on multilingual datasets.

\paragraph{Pivot Translation}~\cite{10.5555/3171837.3171841} This method transforms zero-shot translation into English-centric two-stage translation. The translation model is m-Transformer.

\paragraph{Contrastive Learning (mRASP2 w/o AA)}~\cite{pan-etal-2021-contrastive} This method introduces contrastive learning to learn a universal representation space for the encoder.
For a fair comparison, we do not use the aligned augmentation (AA).

\paragraph{Target Language Prediction (TLP)}~\cite{yang-etal-2021-improving-multilingual} This method leverages a target language prediction task to help the decoder retain information about the target language.

\paragraph{Denoising Training (DT)}~\cite{wang-etal-2021-rethinking-zero} This method introduces a denoising auto-encoder task during training. Specifically, they utilize all English sentences to construct the denoising corpus via text infilling operation.

\paragraph{CrossConST}~\cite{gao-etal-2023-improving} This approach adds KL divergence loss to the Softmax layer to promote consistent cross-lingual representation:
\begin{equation}
    \mathcal{L}_{kl} = \gamma \mathbf{KL}(P_\theta (\mathbf{x},\mathbf{y})||P_\theta (\mathbf{y},\mathbf{y}))
\end{equation}
Where $\gamma$ is the hyperparameter controlling the proportion of $\mathcal{L}_{kl}$.
We set $\gamma=0.25$ on IWSLT2017, $\gamma=0.06$ on OPUS-7 and $\gamma=0.1$ on PC-6. We jointly train $\mathcal{L}_{ce}$ and $\mathcal{L}_{kl}$.

\paragraph{Alpaca}~\cite{alpaca} Alpaca is fine-tuned by Stanford from LLaMA-7B~\cite{touvron2023llama} with 52k instruction data. We reproduce Alpaca using the LoRA~\cite{hu2022lora} setting.

\paragraph{ChatGPT} We use the GPT-3.5-Turbo API\footnote{The API version is GPT-3.5-Turbo-0613.} for translation experiments. Our translation prompt for Alpaca and ChatGPT is reported in Appendix \ref{sec:prompt}.

\subsection{Main Results}

The results of the main experiment are shown in Table \ref{main_result}. For OPUS-7 and PC-6, we report the average BLEU scores for the same target language for convenience, and detailed results for the zero-shot translation directions are shown in Appendix \ref{sec:detailed_result}. The ChrF and COMET results are presented in Appendix \ref{sec:ChrF_and_COMET}. 

The results show that our system significantly improves zero-shot translation performance and maintains supervised translation performance. Specifically, in the supervised translation direction, the performance of our system is comparable to that of m-Transformer, which suggests that our disentangling approach mitigates alignment degradation and maintains the supervised translation performance. In the zero-shot translation direction, our system significantly outperforms all the contrast models, indicating that our additional modeling of linguistic features can assist the model in learning better semantic representation and language generation. Moreover, our method even outperforms the Pivot system on IWSLT2017 and PC-6, which we believe is probably because Pivot accumulates too many errors on these two datasets.

The results of large language models (LLMs) indicate that our task-specific model still has advantages in the era of LLMs. Our model significantly outperforms Alpaca-7B in supervised and zero-shot translation. Compared to the ChatGPT, our model achieves more than 90\% performance with just about 0.1\% of the parameters. Furthermore, our model employs significantly less data and inference is much faster.
\section{Analysis}

In this section, we will analyze what contributes to the performance. We first analyze the effects of each module via an ablation study. Then, we evaluate the off-target rate and visualize the encoder and decoder sentence representations, demonstrating that our approach learns universal representations and generates target language more accurately. Finally, we examine some translation cases to show the usefulness of our approach.

\subsection{Ablation Study}

\begin{table*}[htbp]
    \centering
    \begin{tabular}{c|ccc|cc} \hline
    \multirow{2}*{Model} & \multirow{2}*{Disentangler} & Disentangling & Linguistic & Zero-shot & Supervised \\
    ~ & ~ & Objective & Encoder & Average & Average \\ \hline
    m-Transformer & \ding{56} & \ding{56} & \ding{56} & 17.93 & 34.32 \\
    m-Transformer$^\dag$ & \ding{56} & \ding{56} & \ding{56} & 17.03 & 34.33 \\ \hdashline
    \ding{172} & \ding{52} & \ding{56}  & \ding{56} & 19.88 & 34.39 \\
    \ding{173} & \ding{52} & \ding{52} & \ding{56} & 21.86 & \textbf{34.45} \\
    \ding{174} & \ding{56} & \ding{56} & \ding{52} & 21.27 & 34.34 \\
    \ding{175} & \ding{52} & \ding{56} & \ding{52} & 21.92 & 34.23 \\
    \ding{176} & \ding{52} & \ding{52} & \ding{52} & \textbf{22.70} & 34.25 \\ \hline
    \end{tabular}
    \caption{The results of the ablation study. The markers \ding{52} and \ding{56} indicate the component is involved or not involved, respectively. The m-Transformer$^\dag$ has eight encoder layers and eight decoder layers, similar to our model size.}
    \label{ablation_result}
\end{table*}

To further analyze the effectiveness of our approach, we conducted experiments on different variants of our system on the IWSLT2017. We briefly report the experimental results in Table \ref{ablation_result}.

Compared with the m-Transformer, \ding{174} can significantly improve the zero-shot translation. The gap in zero-shot translation performance remains even when scaling up the m-Transformer (m-Transformer$^\dag$). This indicates that it is effective to leverage the linguistic encoder to learn low-level linguistic features and integrate different levels of representations. We present detailed results and analysis of the scaling experiments in Appendix \ref{sec::scaling} for parameter comparisons.

Compared with \ding{172}, \ding{173} provides a significant improvement in zero-shot translation while maintaining supervised translation performance. This indicates that the disentangling learning task is effective. Disentangling learning can significantly improve the performance of zero-shot translation while mitigating the performance loss in supervised translation caused by semantic alignment. The analysis between \ding{175} and \ding{176} similarly supports this view.

Compared with \ding{173} and \ding{174}, \ding{176} shows that the disentangler and the linguistic encoder can collaborate to improve the zero-shot translation performance.

\begin{figure*}[htbp]
    \centering
    \subfigure[m-Transformer Encoder]{\includegraphics[width=1.5in]{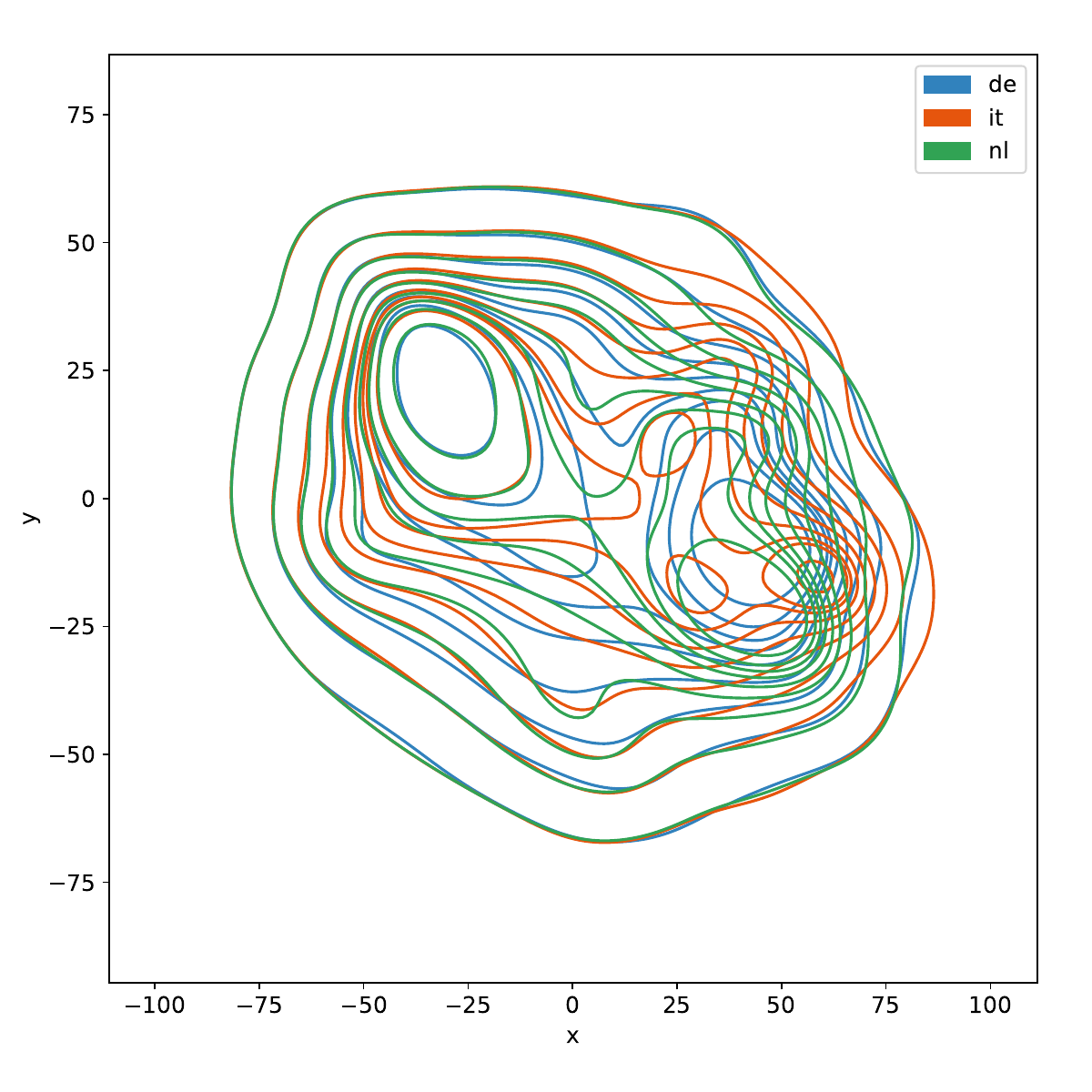}\label{a}}
    \subfigure[mRASP2 w/o AA Encoder]{\includegraphics[width=1.5in]{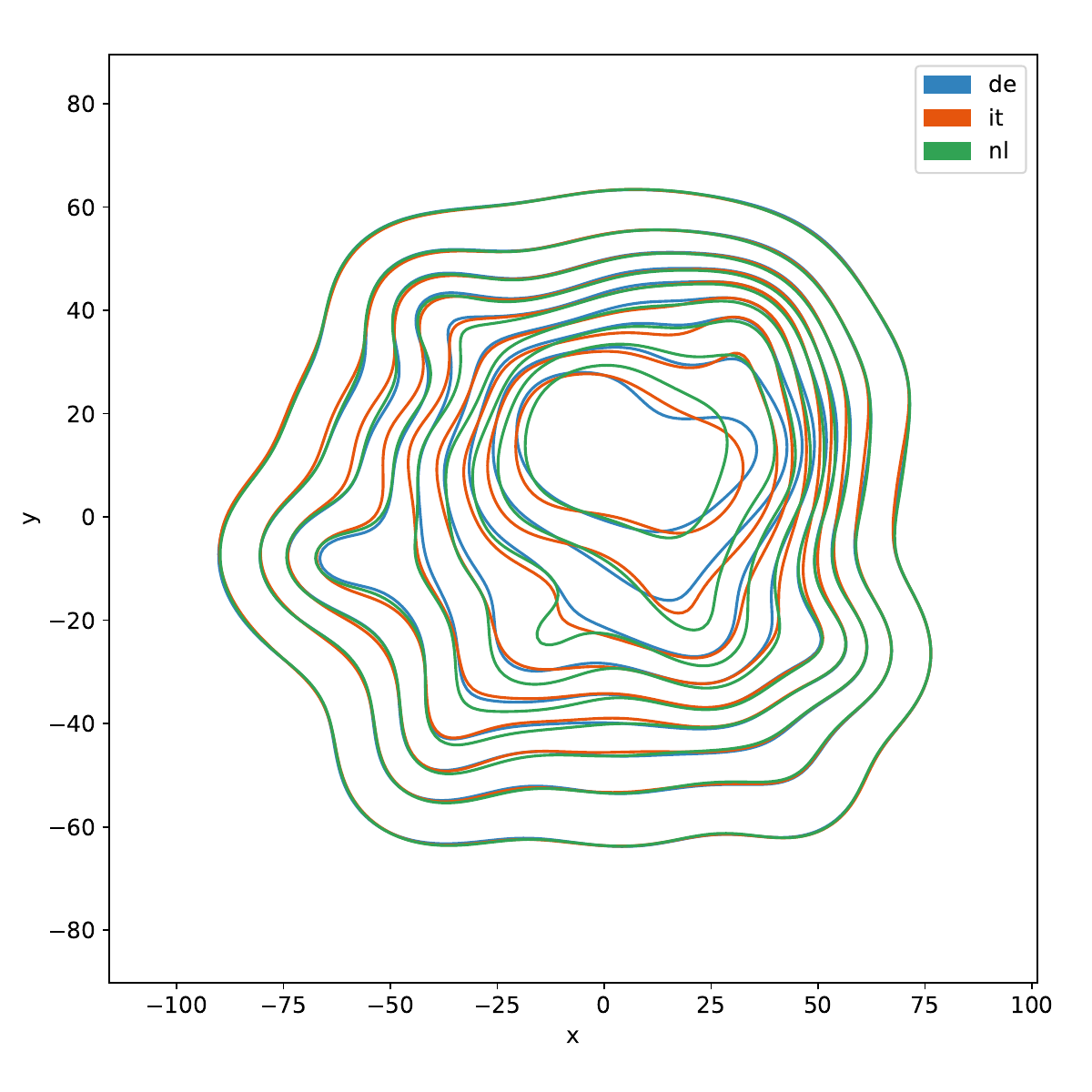}\label{b}}
    \subfigure[Semantic FFN]{\includegraphics[width=1.5in]{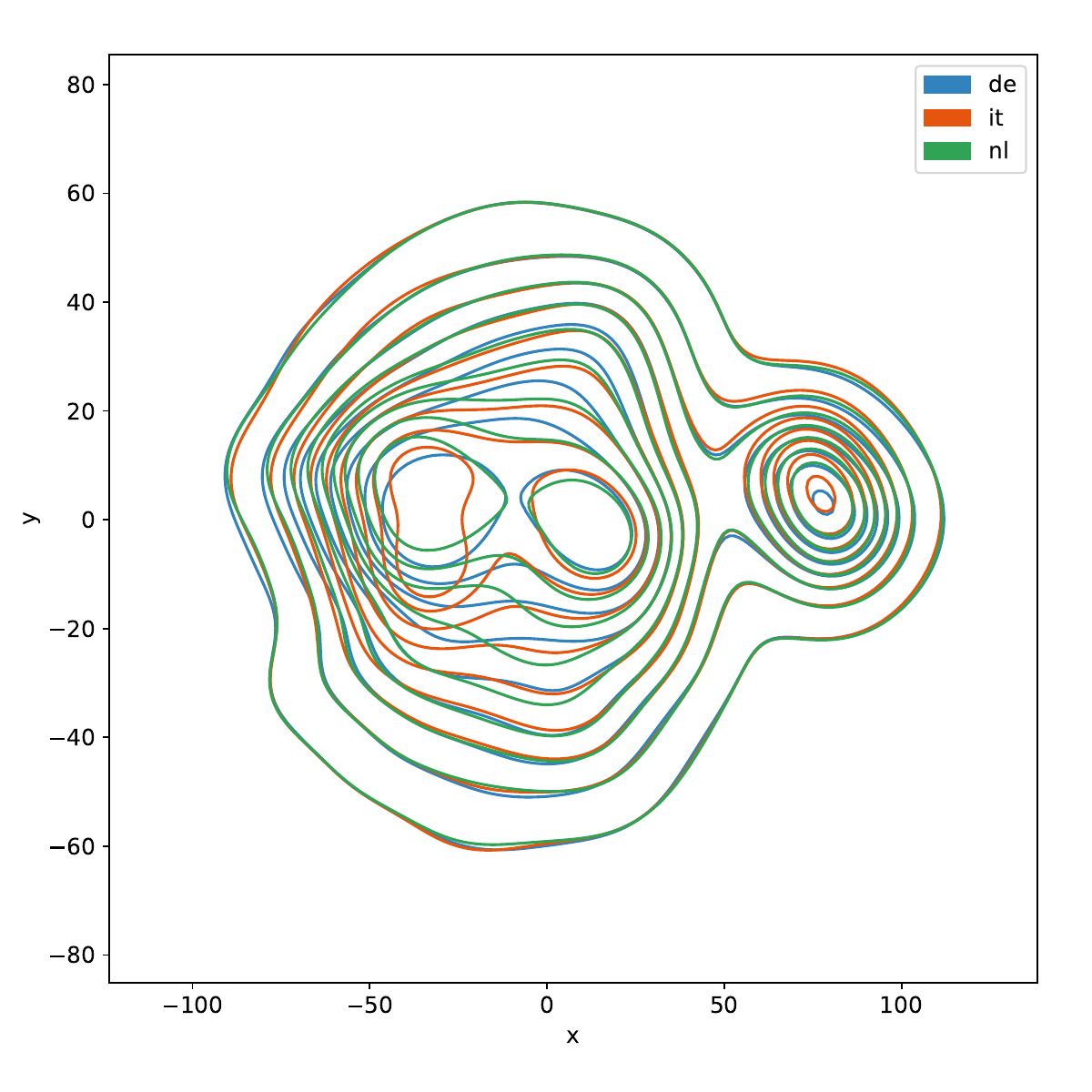}\label{c}}
    \subfigure[Language FFN]{\includegraphics[width=1.5in]{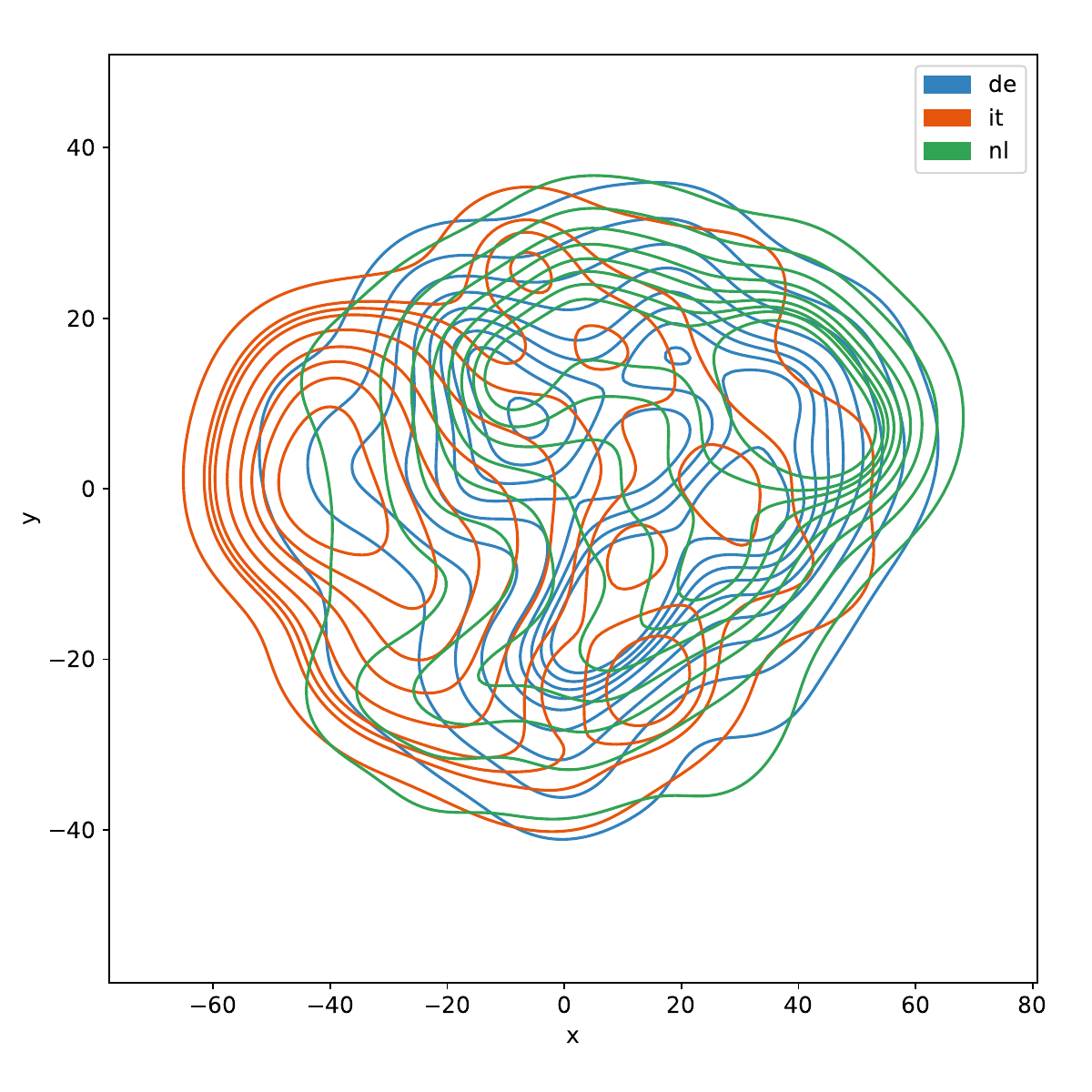}\label{d}} \\
    \subfigure[Decoder Embedding]{\includegraphics[width=1.5in]{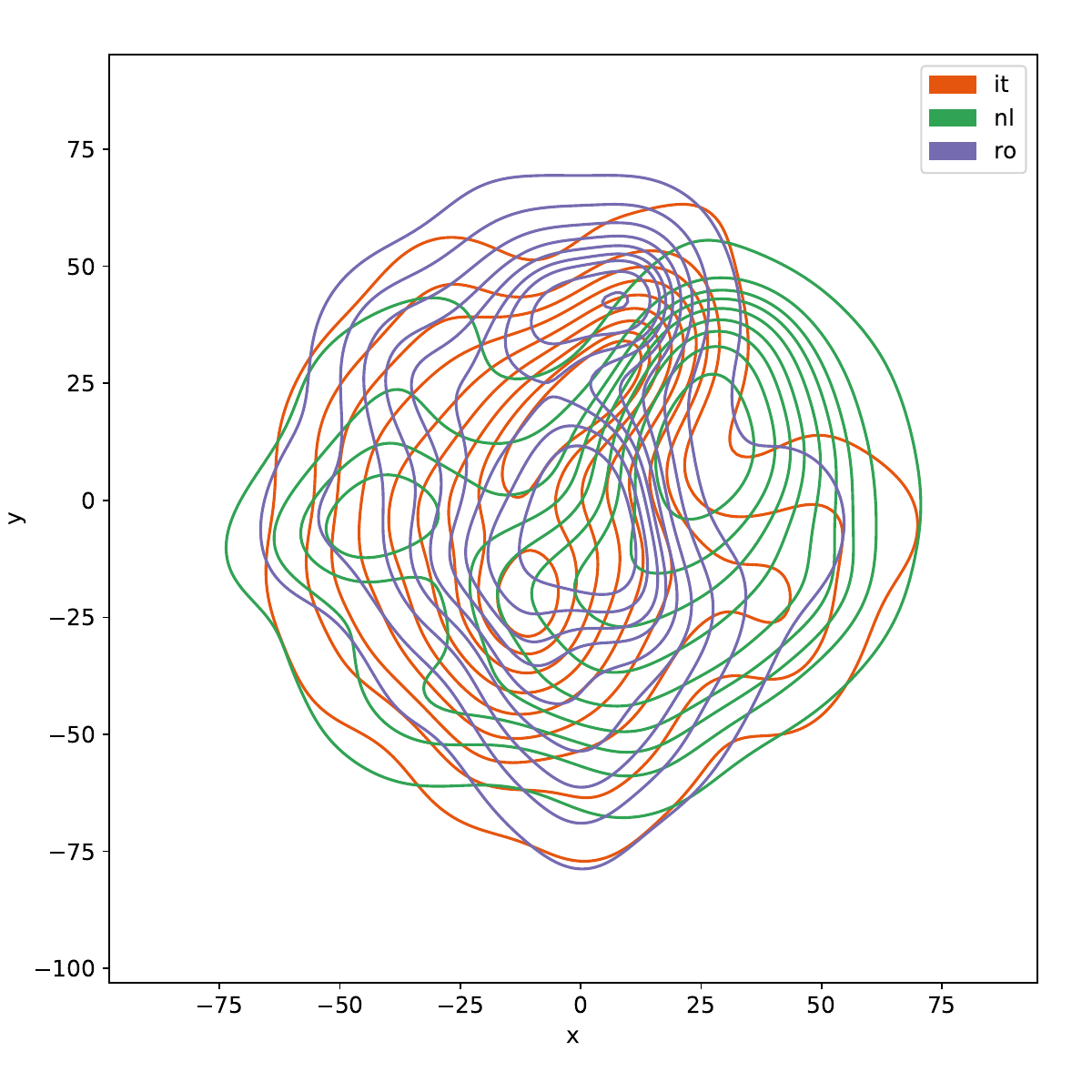}\label{e}}
    \subfigure[Linguistic Encoder]{\includegraphics[width=1.5in]{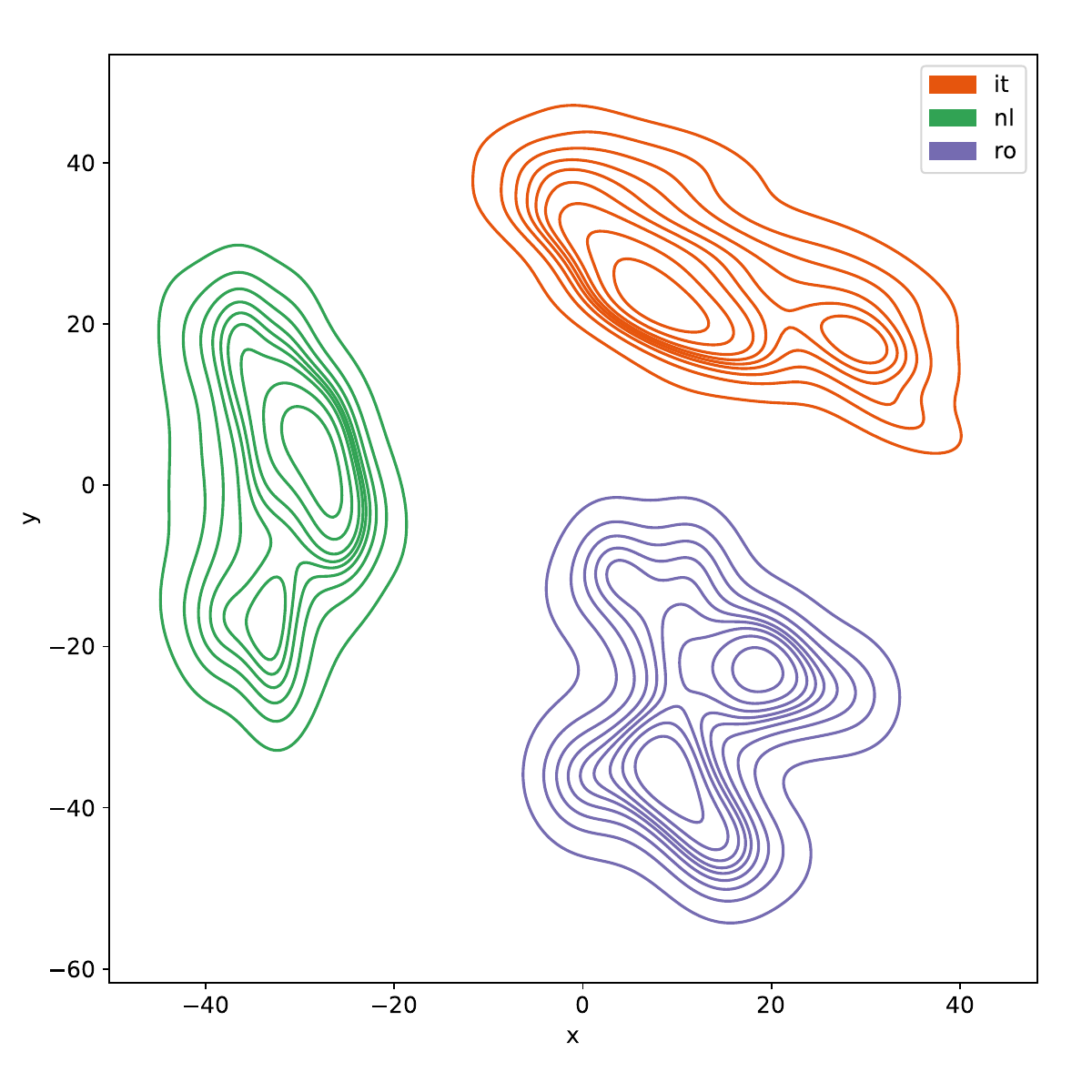}\label{f}}
    \subfigure[m-Transformer Decoder]{\includegraphics[width=1.5in]{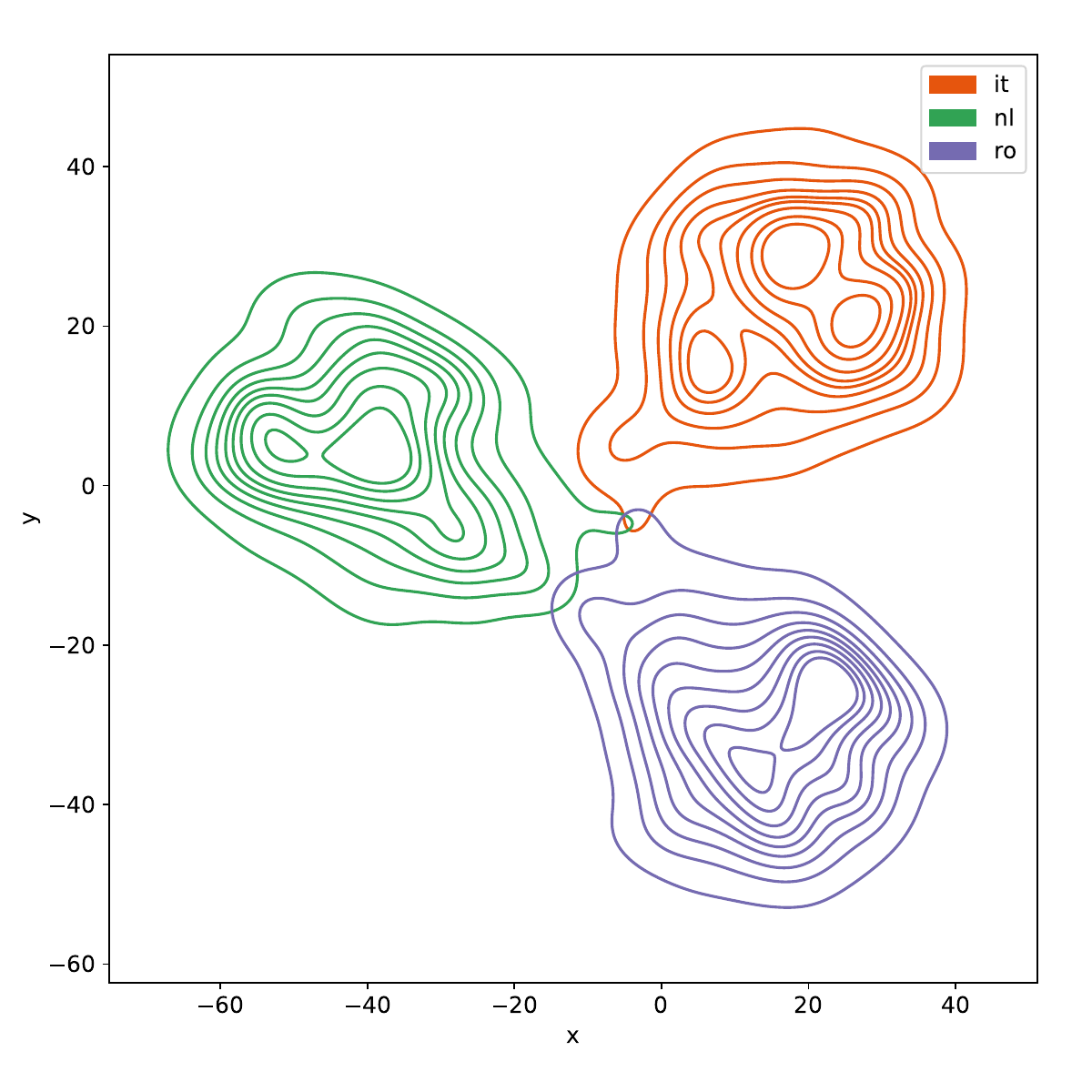}\label{g}}
    \subfigure[Fusion Layer]{\includegraphics[width=1.5in]{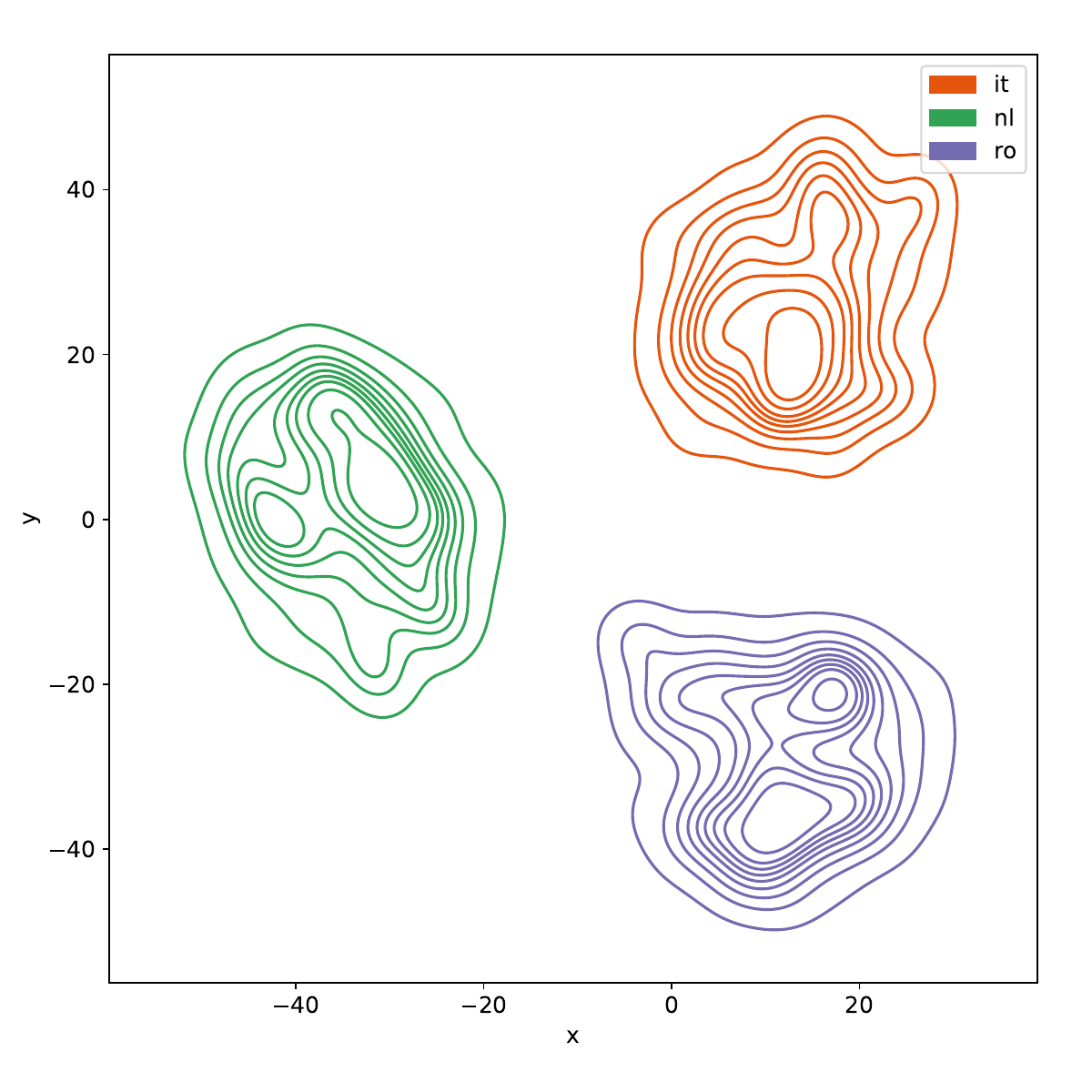}\label{h}}
    \caption{Visualization of the m-Transformer, mRASP2 w/o AA and our system after dimension reduction. The subfigure captions describe the model modules, and we do dimension reduction on the outputs of these modules. The blue line denotes German (De), the orange line denotes Italian (It), the green line denotes Dutch (Nl), and the purple line denotes Romanian (Ro).}
    \label{visualization}
\end{figure*}

\subsection{Off-target Issue}

\begin{table}[t]
    \centering
    \resizebox{\linewidth}{!}{\begin{tabular}{c|ccc} \hline
    ~ & IWSLT2017 & OPUS-7 & PC-6 \\ \hline
    m-Transformer & 6.76\% & 11.58\% & 21.02\%  \\
    mRASP2 w/o AA & 2.84\% & 8.91\% & 8.68\% \\
    CrossConST & 3.07\% & 11.71\% & 7.94\% \\
    Ours & \textbf{2.46\%} & \textbf{8.64\%} & \textbf{7.25\%} \\ \hline
    \end{tabular}}
    \caption{The average off-target rate for the test sets.}
    \label{Off-target}
\end{table}

\begin{table}[t]
    \centering
    \resizebox{\linewidth}{!}{\begin{tabular}{c|ccc} \hline
    ~ & IWSLT2017 & OPUS-7 & PC-6 \\ \hline
    m-Transformer & 18.39 & 15.76 & 7.91  \\
    mRASP2 w/o AA & 22.42 &17.31 & 11.48 \\
    Ours & \textbf{22.69} & \textbf{17.57} & \textbf{11.82} \\ \hline
    \end{tabular}}
    \caption{The average in-target BLEU scores for the test sets.}
    \label{In-target}
\end{table}

The many-to-many multilingual NMT models face the off-target issue~\cite{gu-etal-2019-improved} in zero-shot translation, which seriously impacts the zero-shot translation performance. The off-target issue refers to the problem that the model simply copies the source sentences or generates incorrect target sentences. Related work shows that multilingual NMT models have difficulty establishing mapping relations for the zero-shot translation directions, which may be overfitted to other language directions. To test whether our approach can alleviate the off-target issue, we use \textit{langid} toolkit~\cite{lui-baldwin-2012-langid} to identify the target language and define the off-target rate as the proportion of off-target sentences. We conduct experiments on the contrast systems and our method. Table \ref{Off-target} shows that our method can effectively reduce the off-target rate compared to other systems, proving that our method can provide an effective guide for target language generation.

Furthermore, we test the performance of in-target sentences to demonstrate that our approach improves the in-target translation quality. Considering that there is no comparability between the BLEU scores of different test sets, we extract the part where all model outputs are in-target for testing. Table \ref{In-target} presents the results. Compared with Table \ref{main_result}, the performance gap between different models in the in-target part is smaller, which indicates the importance of mitigating the off-target issue. Meanwhile, the zero-shot performance of our models in the in-target part still significantly outperforms m-Transformer, which implies that our approach achieves better knowledge transfer and improves the zero-shot translation quality.

\subsection{Visualization}

We visualize the sentence representations of the encoder and decoder to verify, respectively, that our disentangler aids semantic alignment and our linguistic encoder achieves more accurate target language generation. We conduct experiments on a subset of IWSLT2017 test sets. We use t-SNE~\cite{van2008visualizing} to reduce the 512-dim representations to 2-dim. 

Figures \ref{a} - \ref{d}: The m-Transformer cannot align different languages well, which is caused by the entanglement of semantic and linguistic features at the representation level. For semantic features, our semantic FFN pulls the semantic representation closer, achieving a similar effect as mRASP2 w/o AA. For linguistic features, our language FFN pushs the different language representations apart, which proves that the language FFN can extract linguistic features.

Figures \ref{e} - \ref{h}: The representation after decoder embedding is language-mixed, while the linguistic encoder can encode different languages into different subspaces, even though no explicit task is designed to guide this. By introducing low-level linguistic features, our system can better distinguish the target language and thus generate the target sentences more accurately.

\subsection{Case Study}

\begin{figure}[t!]
  \centering
  \includegraphics[width=1\linewidth]{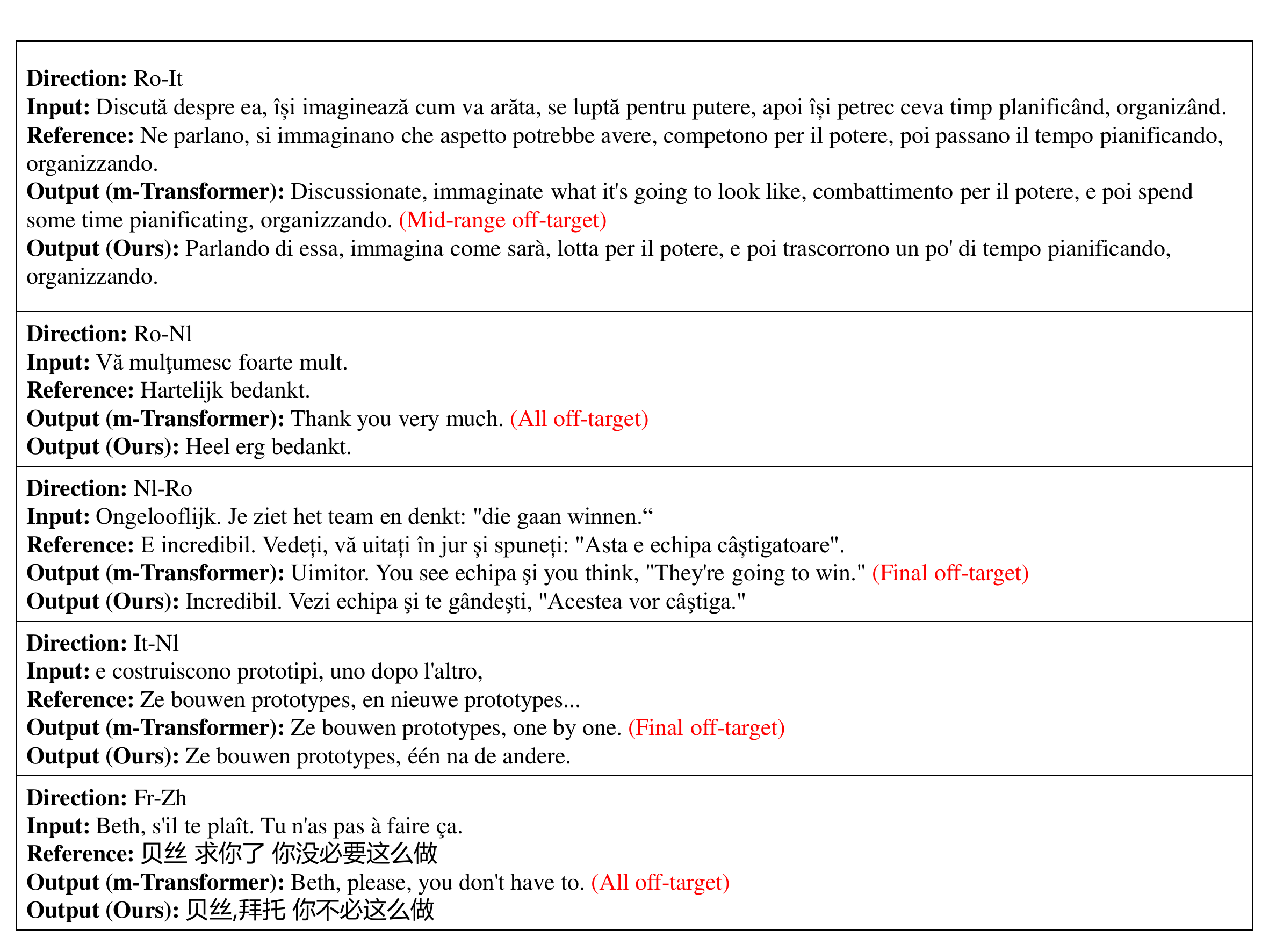}
  \caption{Case study of zero-shot translation. We identify three off-target types which are categorized by off-target position and ratio.}
  \label{case_study}
\end{figure}

To further demonstrate the usefulness of our system, we analyze some translation cases and compare the outputs of our model with that of the m-Transformer in Figure \ref{case_study}. While the m-Transformer translates whole or partial sentences to the wrong language, our model can translate accurately, proving the advantages of our method.

\section{Related Work}

\subsection{Zero-shot Machine Translation}

To improve the zero-shot translation capability of multilingual NMT, researchers have proposed a series of auxiliary training objectives. These works can be divided into two categories.
The first category is designed directly at the encoder so that the encoder outputs language-agnostic representations \citep{pham-etal-2019-improving, DBLP:conf/iclr/WeiW0XYL21}. These works are motivated by the hypothesis that a universal semantic representation can realize better knowledge transfer. \citet{pan-etal-2021-contrastive} propose a contrastive learning task to encourage the model to learn a universal representation space. \citet{gu-feng-2022-improving} use optimal transport theory to close the representation of sentence pairs. However, these works focus only on semantic features and neglect linguistic features, which leads to performance degradation in the supervised translation direction. By integrating both semantic and linguistic features, our approach significantly improves the performance of zero-shot translation while effectively maintaining the supervised direction performance.

The second category is designed at the decoder to provide explicit guidance for target language generation. These works are motivated to improve multilingual representation. \citet{yang-etal-2021-improving-multilingual} propose the target language prediction task at the sequence level and the target gradient regularization at the gradient level. \citet{gao-etal-2023-improving} introduce the KL regularization term for the Softmax layer to promote cross-lingual consistent representation. In contrast to these approaches, we do not explicitly design auxiliary tasks, but utilize a linguistic encoder to fuse low-level representations containing more linguistic features and high-level representations containing more semantic features.

\subsection{Disentangling for Cross-lingual Alignment}

Cross-lingual alignment is a classic problem in multilingual tasks. Although the "pretraining-finetuning" paradigm can improve the cross-lingual ability of the model \citep{liu-etal-2020-multilingual-denoising, xue-etal-2021-mt5}, the entangling of semantics and linguistics can impair the model performance. To address this problem, many researchers have improved it from the perspective of disentangling. \citet{tiyajamorn-etal-2021-language} design auto-encoder to disentangle semantic and linguistic representation for natural language understanding tasks. \citet{yang-etal-2021-simple} remove language identity information from pre-trained multilingual representations by singular value decomposition (SVD) and orthogonal projection. \citet{zhao-etal-2021-inducing} explore three settings for removing language identity signals from multilingual representations, including vector space re-alignment, vector space normalization and input normalization. \citet{wu2022laft} use SVD to extract linguistic features, use language-agnostic semantic information for task-specific training, and re-entangle semantic and linguistic features in the generation phase. We refer to this disentangling idea and utilize disentangling to achieve lossless semantic alignment for multilingual NMT.

\section{Conclusion}

In this paper, we propose to exploit both semantic and linguistic features to improve multilingual NMT. We design a disentangler and disentangling learning task at the encoder side to achieve lossless semantic alignment. Meanwhile, we introduce a linguistic encoder at the decoder side to fuse the low-level representation containing more linguistic features and the high-level representation containing more semantic features to improve the target language generation. Experiments on multilingual datasets indicate that our approach can significantly improve the performance of zero-shot translation while effectively maintaining the supervised translation performance. The analysis further demonstrates that utilizing both semantic and linguistic features can help the model learn semantic representation non-destructively and achieve more accurate target language generation.
\section*{Limitations}

Although our approach can significantly improve zero-shot translation performance while maintaining supervised translation performance, the performance improvement in the supervised direction is limited. On the one hand, the datasets of our experiments contain only several languages, and the baseline model can already model the supervised translation well. On the other hand, our approach to modeling linguistic features is still preliminary and implicit. Future work could consider using larger multilingual datasets as well as designing better methods to model linguistic features.
\section*{Acknowledgments}

We thank all the anonymous reviewers for their insightful and valuable comments. This work was supported by a grant from the National Natural Science Foundation of China (No. 62376260).

\bibliography{anthology,custom}

\newpage

\appendix





\section{Statistics of the Datasets}
\label{sec:dataset_description}

Table \ref{dataset_description} summarizes information about the multilingual training datasets.

\begin{table}[htbp]
\linespread{1.5}
\centering
\scalebox{0.86}{\begin{tabular}{ccc}
\hline
\textbf{Dataset} & \textbf{Language Pairs} & \textbf{Size}\\ \hline
IWSLT2017 & En $\leftrightarrow$ \{De, It, Nl, Ro\} & 1.8M \\
OPUS-7  & En $\leftrightarrow$ \{Ar, De, Fr, Nl, Ru, Zh\} & 12.0M \\
PC-6 & En $\leftrightarrow$ \{Cs, Kk, Ro, Ru, Tr\} & 7.9M \\ \hline
\end{tabular}}
\caption{A brief description of the datasets.}
\label{dataset_description}
\end{table}

\section{Translation Prompt for Alpaca and ChatGPT}
\label{sec:prompt}

For Alpaca, we follow the official template. We set \textit{Instruction} to \textit{"Translate input from <src> to <tgt>."} and set \textit{Input} to the source sentence.
For ChatGPT, we use the following format to structure the inputs: \textit{"Translate the sentences from <src> to <tgt>: <input>."}. \textit{<src>} denotes the source language, \textit{<tgt>} denotes the target language and \textit{<input>} denotes the source sentence.

\section{Detailed Results on the OPUS-7 and PC-6 Test Sets}
\label{sec:detailed_result}

The detailed results on the OPUS-7 and PC-6 test sets are shown in Table \ref{Opus-7-detail} and Table \ref{PC-6-detail} respectively.

\section{ChrF and COMET Metrics}
\label{sec:ChrF_and_COMET}

For the ChrF and COMET metrics, we evaluate the m-Transformer, mRASP2 w/o AA and our model. The results are presented in Table \ref{ChrF} and Table \ref{COMET} respectively.

\section{Scaling Experiments on the IWSLT2017 Test Set}
\label{sec::scaling}

Considering that our approach introduces additional parameters, we scale the model sizes of the m-Transformer and mRASP2 w/o AA for a fair comparison. As shown in Table \ref{scaling}, our model outperforms these variant models. We note a significant performance degradation of the m-Transformer when the model size reaches 100.4M. We attribute this to the limited size of the IWSLT2017 training set, thus the model performance largely depends on the method, rather than the model size. This further demonstrates the effectiveness of our method. 



\begin{table}[t]
    \centering
    
    \resizebox{\linewidth}{!}{\begin{tabular}{c|cccccc|c}
    m-Transformer & Ar & De & Fr & Nl & Ru & Zh & x (avg) \\ \hline
    Ar & - & 12.13 & 24.96 & 14.33 & 21.31 & 27.12 & 19.97 \\
    De & 5.74 & - & 19.42 & 21.15 & 13.26 & 5.55 & 13.02 \\ 
    Fr & 16.80 & 17.97 & - & 21.57 & 21.77 & 22.46 & 20.11 \\
    Nl & 4.88 & 20.02 & 22.60 & - & 10.69 & 2.78 & 12.19 \\ 
    Ru & 18.72 & 13.58 & 25.36 & 13.52 & - & 24.30 & 19.09 \\
    Zh & 17.50 & 9.23 & 23.22 & 8.51 & 19.18 & - & 15.53 \\ \hline
    x (avg) & 12.73 & 14.58 & 23.11 & 15.82 & 17.24 & 16.44 & 16.65 \\ \hline
    \end{tabular}}
    
    \vspace{1em}

    \resizebox{\linewidth}{!}{\begin{tabular}{c|cccccc|c}
    Pivot & Ar & De & Fr & Nl & Ru & Zh & x (avg) \\ \hline
    Ar & - & 15.67  & 29.81  & 18.16  & 26.00  & 22.67  & 22.46 \\
    De & 7.30  & - & 23.07  & 23.16  & 16.02  & 5.13  & 14.94 \\
    Fr & 20.32  & 20.90  & - & 23.80  & 25.01  & 25.20  & 23.05 \\
    Nl & 6.55  & 21.40  & 25.47  & - & 13.40  & 3.34  & 14.03 \\
    Ru & 22.56  & 16.72  & 30.56  & 16.27  & - & 22.21  & 21.66 \\
    Zh & 22.23  & 12.70  & 30.13  & 11.10  & 24.47  & - & 20.13 \\ \hline
    x (avg) & 15.79  & 17.48  & 27.81  & 18.50  & 20.98  & 15.71  & 19.38 \\ \hline
    \end{tabular}}
    
    \vspace{1em}
    
    \resizebox{\linewidth}{!}{\begin{tabular}{c|cccccc|c}
    mRASP2 w/o AA & Ar & De & Fr & Nl & Ru & Zh & x (avg) \\ \hline
    Ar & - & 14.74 & 27.06 & 15.71 & 22.86 & 26.05 & 21.28 \\
    De & 6.48 & - & 22.09 & 23.06 & 14.91 & 6.82 & 14.67 \\ 
    Fr & 19.00 & 20.01 & - & 23.85 & 23.89 & 26.30 & 22.61 \\
    Nl & 6.08 & 21.75 & 24.86 & - & 12.83 & 3.90 & 13.89 \\ 
    Ru & 20.02 & 15.51 & 28.10 & 14.84 & - & 25.52 & 20.80 \\
    Zh & 19.43 & 10.67 & 26.22 & 9.82 & 21.97 & - & 17.62 \\ \hline
    x (avg) & 14.20 & 16.54 & 25.67 & 17.46 & 19.29 & 17.72 & 18.48 \\ \hline
    \end{tabular}}

    \vspace{1em}
    
    \resizebox{\linewidth}{!}{\begin{tabular}{c|cccccc|c}
    TLP & Ar & De & Fr & Nl & Ru & Zh & x (avg) \\ \hline
    Ar & - & 11.59 & 23.91 & 13.91 & 20.91 & 25.03 & 19.07 \\
    De & 6.24 & - & 19.72 & 21.40 & 13.29 & 4.60 & 13.05 \\
    Fr & 17.52 & 17.47 & - & 21.64 & 22.22 & 24.81 & 20.73 \\
    Nl & 4.93 & 19.87 & 22.47 & - & 10.53 & 2.96 & 12.15 \\
    Ru & 19.18 & 13.60 & 24.90 & 13.52 & - & 24.05 & 19.05 \\
    Zh & 17.65 & 9.08 & 22.56 & 8.65 & 19.52 & - & 15.49 \\ \hline
    x (avg) & 13.10 & 14.32 & 22.71 & 15.82 & 17.29 & 16.29 & 16.59 \\ \hline
    \end{tabular}}

    \vspace{1em}
    
    \resizebox{\linewidth}{!}{\begin{tabular}{c|cccccc|c}
    DT & Ar & De & Fr & Nl & Ru & Zh & x (avg) \\ \hline
    Ar & - & 12.80 & 25.81 & 14.77 & 22.27 & 28.52 & 20.83 \\
    De & 6.76 & - & 20.62 & 22.01 & 13.88 & 6.01 & 13.85 \\ 
    Fr & 18.08 & 18.11 & - & 21.98 & 23.08 & 25.47 & 21.34 \\
    Nl & 5.22 & 20.83 & 23.31 & - & 11.52 & 3.33 & 12.84 \\
    Ru & 20.13 & 14.06 & 26.21 & 13.78 & - & 25.93 & 20.02 \\
    Zh & 19.00 & 9.47 & 23.79 & 8.68 & 20.71 & - & 16.33 \\ \hline
    x (avg) & 13.84 & 15.06 & 23.95 & 16.24 & 18.29 & 17.85 & 17.54 \\ \hline
    \end{tabular}}
    
    \vspace{1em}

    \resizebox{\linewidth}{!}{\begin{tabular}{c|cccccc|c}
    CrossConST & Ar & De & Fr & Nl & Ru & Zh & x (avg) \\ \hline
    Ar & - & 13.02 & 25.88 & 14.95 & 21.70 & 25.48 & 20.21 \\
    De & 6.30 & - & 19.68 & 21.94 & 13.74 & 5.09 & 13.35 \\ 
    Fr & 18.09 & 18.22 & - & 22.56 & 21.91 & 23.17 & 20.79 \\
    Nl & 5.26 & 20.45 & 22.96 & - & 11.28 & 2.51 & 12.49 \\
    Ru & 19.76 & 14.18 & 26.17 & 14.12 & - & 23.89 & 19.62 \\
    Zh & 19.01 & 9.35 & 23.67 & 8.92 & 19.70 & - & 16.13 \\ \hline
    x (avg) & 13.68 & 15.04 & 23.67 & 16.50 & 17.67 & 16.03 & 17.10 \\ \hline
    \end{tabular}}
    
    \vspace{1em}
    
    \resizebox{\linewidth}{!}{\begin{tabular}{c|cccccc|c}
    Ours & Ar & De & Fr & Nl & Ru & Zh & x (avg) \\ \hline
    Ar & - & 14.60  & 27.78  & 15.41  & 24.50  & 26.49  & 21.75 \\
    De & 7.04  & - & 22.31  & 23.04  & 14.84  & 6.37  & 14.72 \\
    Fr & 19.89  & 20.00  & - & 23.90  & 24.92  & 25.10  & 22.76 \\
    Nl & 6.06  & 21.61  & 24.87  & - & 12.81  & 3.58  & 13.79 \\
    Ru & 20.84  & 15.85  & 28.66  & 14.91  & - & 25.46  & 21.15 \\
    Zh & 20.11  & 11.14  & 27.38  & 9.83  & 23.48  & - & 18.39 \\ \hline
    x (avg) & 14.79  & 16.64  & 26.20  & 17.42  & 20.11  & 17.40  & 18.76 \\ \hline
    \end{tabular}}

    \caption{The performance of the contrast models and our model on the OPUS-7 dataset.}

    \label{Opus-7-detail}

\end{table}

\begin{table}[t]
    \centering
    
    \resizebox{\linewidth}{!}{\begin{tabular}{c|ccccc|c}
    m-Transformer & Cs & Kk & Ro & Ru & Tr & x (avg) \\ \hline
    Cs & - & 1.54  & 14.63  & 12.93  & 6.16  & 8.81 \\
    Kk & 1.73  & - & 2.61  & 10.24  & 2.99  & 4.40 \\
    Ro & 11.11  & 1.87  & - & 12.98  & 7.25  & 8.31 \\
    Ru & 10.53  & 11.10  & 14.76  & - & 5.40  & 10.45 \\
    Tr & 8.80  & 2.57  & 12.21  & 7.47  & - & 7.76 \\ \hline
    x (avg) & 8.05  & 4.27  & 11.05  & 10.91  & 5.45  & 7.94 \\ \hline
    \end{tabular}}
    
    \vspace{1em}

    \resizebox{\linewidth}{!}{\begin{tabular}{c|ccccc|c}
    Pivot & Cs & Kk & Ro & Ru & Tr & x (avg) \\ \hline
    Cs & - & 1.01 & 21.4 & 20.08 & 10.19 & 13.17 \\
    Kk & 1.49 & - & 2.09 & 5.07 & 2.29 & 2.74 \\
    Ro & 21.1 & 1.32 & - & 24.1 & 11.43 & 14.49 \\
    Ru & 19.08 & 4.99 & 23.42 & - & 9.27 & 14.19 \\
    Tr & 13.23 & 1.19 & 16.74 & 11.54 & - & 10.68 \\ \hline
    x (avg) & 13.73  & 2.13  & 15.91  & 15.20  & 8.30  & 11.05 \\ \hline
    \end{tabular}}
    
    \vspace{1em}

    \resizebox{\linewidth}{!}{\begin{tabular}{c|ccccc|c}
    mRASP2 w/o AA & Cs & Kk & Ro & Ru & Tr & x (avg) \\ \hline
    Cs & - & 1.11  & 20.70  & 22.09  & 9.94  & 13.46 \\
    Kk & 2.01  & - & 2.35  & 7.65  & 2.33  & 3.59 \\
    Ro & 20.32  & 1.60  & - & 24.42  & 11.94  & 14.57 \\
    Ru & 21.88  & 6.53  & 23.51  & - & 8.65  & 15.15 \\
    Tr & 12.86  & 2.16  & 16.09  & 10.31  & - & 10.36 \\ \hline
    x (avg) & 14.27  & 2.85  & 15.67  & 16.12  & 8.22  & 11.42 \\ \hline
    \end{tabular}}

    \vspace{1em}

    \resizebox{\linewidth}{!}{\begin{tabular}{c|ccccc|c}
    TLP & Cs & Kk & Ro & Ru & Tr & x (avg) \\ \hline
    Cs & - & 1.37  & 14.50  & 12.75  & 7.14  & 8.94 \\
    Kk & 1.96  & - & 2.53  & 10.92  & 3.06  & 4.62 \\
    Ro & 13.36  & 2.07  & - & 12.25  & 7.88  & 8.89 \\
    Ru & 13.71  & 10.83  & 15.96  & - & 6.58  & 11.77 \\
    Tr & 9.66  & 2.82  & 12.41  & 7.25  & - & 8.03 \\ \hline
    x (avg) & 9.67  & 4.27  & 11.35  & 10.79  & 6.17  & 8.45 \\ \hline
    \end{tabular}}
    
    \vspace{1em}

    \resizebox{\linewidth}{!}{\begin{tabular}{c|ccccc|c}
    DT & Cs & Kk & Ro & Ru & Tr & x (avg) \\ \hline
    Cs & - & 1.53  & 19.03  & 18.83  & 8.21  & 11.90 \\
    Kk & 2.13  & - & 2.81  & 12.43  & 3.58  & 5.24 \\
    Ro & 17.34  & 2.27  & - & 20.63  & 10.99  & 12.81 \\
    Ru & 18.75  & 12.23  & 19.51  & - & 8.35  & 14.71 \\
    Tr & 12.06  & 3.35  & 15.95  & 10.64  & - & 10.50 \\ \hline
    x (avg) & 12.57  & 4.85  & 14.32  & 15.63  & 7.78  & 11.03 \\ \hline
    \end{tabular}}
    
    \vspace{1em}

    \resizebox{\linewidth}{!}{\begin{tabular}{c|ccccc|c}
    CrossConST & Cs & Kk & Ro & Ru & Tr & x (avg) \\ \hline
    Cs & - & 0.89  & 20.44  & 18.13  & 9.50  & 12.24 \\
    Kk & 1.75  & - & 1.93  & 5.20  & 1.82  & 2.68 \\
    Ro & 19.96  & 1.19  & - & 20.98  & 10.48  & 13.15 \\
    Ru & 18.45  & 5.23  & 21.08  & - & 7.84  & 13.15 \\
    Tr & 11.95  & 1.86  & 14.80  & 9.34  & - & 9.49 \\ \hline
    x (avg) & 13.03  & 2.29  & 14.56  & 13.41  & 7.41  & 10.14 \\ \hline
    \end{tabular}}
    
    \vspace{1em}

    \resizebox{\linewidth}{!}{\begin{tabular}{c|ccccc|c}
    Ours & Cs & Kk & Ro & Ru & Tr & x (avg) \\ \hline
    Cs & - & 1.37  & 22.11  & 21.63  & 10.93  & 14.01 \\
    Kk & 1.87  & - & 2.45  & 9.48  & 2.73  & 4.13 \\
    Ro & 20.22  & 1.91  & - & 24.05  & 13.94  & 15.03 \\
    Ru & 20.66  & 11.03  & 23.94  & - & 10.15  & 16.45 \\
    Tr & 13.76  & 2.50  & 19.18  & 11.70  & - & 11.79 \\ \hline
    x (avg) & 14.13  & 4.20  & 16.92  & 16.72  & 9.44  & 12.28 \\ \hline
    \end{tabular}}
    
    \caption{The performance of the contrast models and our model on the PC-6 dataset.}

    \label{PC-6-detail}

\end{table}


\begin{table*}[t]
    \centering
    
    \resizebox{\linewidth}{!}{\begin{tabular}{c|ccccccc|c} \hline
    \multirow{2}*{IWSLT2017} & \multirow{2}*{De $\leftrightarrow$ It} & \multirow{2}*{De $\leftrightarrow$ Nl} & \multirow{2}*{De $\leftrightarrow$ Ro} & \multirow{2}*{It $\leftrightarrow$ Nl} & \multirow{2}*{It $\leftrightarrow$ Ro} &\multirow{2}*{Nl $\leftrightarrow$ Ro} & Zero-shot & Supervised \\
    ~ & ~ & ~ & ~ & ~ & ~ & ~ & Average & Average \\ \hline
    m-Transformer & 43.09 & 46.50 & 42.65 & 43.63 & 43.72 & 43.53 & 43.85 & 58.39 \\
    mRASP2 w/o AA & 48.36 & 50.93 & 47.59 & 49.42 & 49.54 & 48.86 & 49.11 & 58.47 \\
    Ours & 48.83 & 51.12 & 47.97 & 49.63 & 49.90 & 49.11 & 49.43 & 58.46 \\ \hline
    \end{tabular}}
    
    \vspace{0.5em}
    
    \resizebox{\linewidth}{!}{\begin{tabular}{c|ccccccc|c} \hline
    \multirow{2}*{OPUS-7} & \multirow{2}*{x $\rightarrow$ Ar} & \multirow{2}*{x $\rightarrow$ De} & \multirow{2}*{x $\rightarrow$ Fr} & \multirow{2}*{x $\rightarrow$ Nl} & \multirow{2}*{x $\rightarrow$ Ru} &\multirow{2}*{x $\rightarrow$ Zh} & Zero-shot & Supervised \\
    ~ & ~ & ~ & ~ & ~ & ~ & ~ & Average & Average \\ \hline
    m-Transformer & 37.69 & 37.08 & 48.16 & 37.60 & 41.00 & 25.22 & 37.79 & 53.77 \\
    mRASP2 w/o AA & 39.76 & 40.55 & 50.83 & 39.68 & 43.78 & 26.03 & 40.10 & 53.28 \\
    Ours & 40.40 & 40.67 & 51.34 & 39.55 & 44.53 & 26.95 & 40.57 & 53.86 \\ \hline
    \end{tabular}}
    
    \vspace{0.5em}
    
    \resizebox{\linewidth}{!}{\begin{tabular}{c|cccccc|c} \hline
    \multirow{2}*{PC-6} & \multirow{2}*{x $\rightarrow$ Cs} & \multirow{2}*{x $\rightarrow$ Kk} & \multirow{2}*{x $\rightarrow$ Ro} & \multirow{2}*{x $\rightarrow$ Ru} & \multirow{2}*{x $\rightarrow$ Tr} & Zero-shot & Supervised \\
    ~ & ~ & ~ & ~ & ~ & ~ & Average & Average \\ \hline
    m-Transformer & 30.82 & 16.66 & 36.03 & 30.38 & 28.07 & 28.39 & 49.66 \\
    mRASP2 w/o AA & 38.71 & 17.25 & 42.02 & 41.69 & 35.65 & 35.06 & 49.64 \\
    Ours & 38.84 & 19.44 & 43.41 & 43.17 & 37.17 & 36.40 & 49.80 \\ \hline
    \end{tabular}}
    \caption{The ChrF scores on the IWSLT2017, OPUS-7 and PC-6 test sets.}
    \label{ChrF}
\end{table*}

\begin{table*}[t]
    \centering
    
    \resizebox{\linewidth}{!}{\begin{tabular}{c|ccccccc|c} \hline
    \multirow{2}*{IWSLT2017} & \multirow{2}*{De $\leftrightarrow$ It} & \multirow{2}*{De $\leftrightarrow$ Nl} & \multirow{2}*{De $\leftrightarrow$ Ro} & \multirow{2}*{It $\leftrightarrow$ Nl} & \multirow{2}*{It $\leftrightarrow$ Ro} &\multirow{2}*{Nl $\leftrightarrow$ Ro} & Zero-shot & Supervised \\
    ~ & ~ & ~ & ~ & ~ & ~ & ~ & Average & Average \\ \hline
    m-Transformer & 70.14 & 74.01 & 71.54 & 71.73 & 74.07 & 73.59 & 72.51 & 83.82 \\
    mRASP2 w/o AA & 77.01 & 79.27 & 78.14 & 78.49 & 81.24 & 80.16 & 79.05 & 83.80 \\
    Ours & 77.21 & 79.45 & 78.19 & 78.60 & 81.33 & 80.11 & 79.15 & 83.83 \\ \hline
    \end{tabular}}
    
    \vspace{0.5em}
    
    \resizebox{\linewidth}{!}{\begin{tabular}{c|ccccccc|c} \hline
    \multirow{2}*{OPUS-7} & \multirow{2}*{x $\rightarrow$ Ar} & \multirow{2}*{x $\rightarrow$ De} & \multirow{2}*{x $\rightarrow$ Fr} & \multirow{2}*{x $\rightarrow$ Nl} & \multirow{2}*{x $\rightarrow$ Ru} &\multirow{2}*{x $\rightarrow$ Zh} & Zero-shot & Supervised \\
    ~ & ~ & ~ & ~ & ~ & ~ & ~ & Average & Average \\ \hline
    m-Transformer & 72.33 & 65.77 & 70.67 & 68.97 & 73.48 & 71.52 & 70.46 & 79.45 \\
    mRASP2 w/o AA & 74.36 & 68.41 & 72.77 & 70.89 & 76.15 & 74.02 & 72.77 & 79.25 \\
    Ours & 74.64 & 68.74 & 73.33 & 71.02 & 76.65 & 74.18 & 73.09 & 79.66 \\ \hline
    \end{tabular}}
    
    \vspace{0.5em}
    
    \resizebox{\linewidth}{!}{\begin{tabular}{c|cccccc|c} \hline
    \multirow{2}*{PC-6} & \multirow{2}*{x $\rightarrow$ Cs} & \multirow{2}*{x $\rightarrow$ Kk} & \multirow{2}*{x $\rightarrow$ Ro} & \multirow{2}*{x $\rightarrow$ Ru} & \multirow{2}*{x $\rightarrow$ Tr} & Zero-shot & Supervised \\
    ~ & ~ & ~ & ~ & ~ & ~ & Average & Average \\ \hline
    m-Transformer & 58.10 & 46.51 & 62.63 & 62.25 & 54.84 & 56.86 & 76.36 \\
    mRASP2 w/o AA & 68.98 & 50.74 & 69.19 & 70.95 & 64.85 & 64.94 & 76.32 \\
    Ours & 70.13 & 52.99 & 72.32 & 72.98 & 66.86 & 67.05 & 77.31 \\ \hline
    \end{tabular}}
    \caption{The COMET scores on the IWSLT2017, OPUS-7 and PC-6 test sets.}
    \label{COMET}
\end{table*}


\begin{table*}[t]
    \centering
    
    \begin{tabular}{c|ccccccc|c} \hline
    \multirow{2}*{Models} & \multirow{2}*{$l_{enc}$} & \multirow{2}*{$l_{dec}$} & \multirow{2}*{Model Size} & Zero-shot & Supervised \\
    ~ & ~ & ~ & ~ & Average & Average \\ \hline
    m-Transformer & 6 & 6 & 79.4M & 17.93 & \textbf{34.32} \\
    m-Transformer$^\dag$ & 6 & 8 & 87.8M & 18.10 & 34.23 \\
    m-Transformer$^\ddag$ & 8 & 8 & 100.4M & 16.84 & 34.21 \\
    mRASP2 w/o AA & 6 & 6 & 79.4M & 22.41 & 34.06 \\
    mRASP2 w/o AA$^\dag$ & 8 & 8 & 100.4M & 22.46 & 34.19 \\
    Ours & 6 & 6 & 99.4M & \textbf{22.70} & 34.25 \\ \hline
    \end{tabular}
    
    \caption{The scaling experiments on the IWSLT2017 test set. $l_{enc}$ and $l_{dec}$ denote layers of encoder and decoder respectively. The m-Transformer$^\dag$, m-Transformer$^\ddag$ and mRASP2 w/o AA$^\dag$ denote scaled variants of the original model.}
    \label{scaling}
\end{table*}


\end{document}